\pdfoutput=1
\documentclass[runningheads]{llncs}

\usepackage[T1]{fontenc}
\usepackage[dvipsnames]{xcolor}
\usepackage{amsmath}
\usepackage{amssymb}
\usepackage{cite}
\usepackage{xspace}
\usepackage[hyphens]{url}
\usepackage[labelfont=bf,font=small,tableposition=bottom]{caption}
\usepackage[skip=3pt]{subcaption}
\newcommand{\eg}{\emph{e.g}.\xspace}

\usepackage{graphicx}
\graphicspath{{figures/qual/}{./}}
\usepackage{booktabs}
\usepackage{enumitem}
\usepackage{algorithm}
\usepackage{algpseudocode}
\usepackage[accsupp]{axessibility}

\usepackage{hyperref}

\newcommand{\qualbaseline}{Baseline}

\begin{document}

\title{Multi-View Transformer with Geometry-Guided Multi-Modal ControlNet\\
for 3D Virtual Try-On}

\titlerunning{GS-VTON-MVT: Geometry-Guided Multi-View 3D Virtual Try-On}

\author{Sarthak Batra \and Armin Mustafa \and Adrian Hilton}

\authorrunning{S.~Batra et al.}

\institute{University of Surrey, Guildford, UK\\
\email{\{s.batra,armin.mustafa,a.hilton\}@surrey.ac.uk}}

\maketitle

\begin{abstract}
Editing a 3D Gaussian Splatting (3DGS) capture of a person so that they wear a new garment requires every rendered view to agree on the same garment details. Existing 3D virtual try-on (VTON) pipelines, however, edit training views largely independently and fuse them with hand-designed heuristics such as manually chosen reference views and fixed attention-blending weights. The resulting inconsistencies are baked into the 3D representation and surface as texture drift, blurred prints, and implausible geometry in novel views.
We present \textbf{GS-VTON-MVT}, a three-stage framework that combines \emph{implicit} and \emph{explicit} 3D reasoning for consistent 3DGS try-on.
(i) A \emph{multi-view inpainting transformer} jointly edits a set of training views with full cross-view self-attention, so that cross-view correspondence is learned rather than imposed.
(ii) A \emph{geometry-guided multi-modal ControlNet} conditions the diffusion editor on the garment image together with depth and surface normals rendered from the current 3DGS state, supplying explicit structure where visual evidence is weakest.
(iii) A \emph{pose-adaptive persona-aware attention} replaces the fixed blending weight between current-view and reference-view features with weights derived from camera-pose proximity.
On the 3D-VTONBench benchmark, GS-VTON-MVT improves FID, KID, LPIPS-based cross-view consistency, and user preference over general-purpose 3D editors and prior 3D try-on methods.
\end{abstract}

\section{Introduction}
\label{sec:intro}

Virtual try-on (VTON) lets a user see how a garment would look on their own body without a physical fitting. Beyond e-commerce, it is increasingly relevant to AR/VR avatars and telepresence, where a garment must remain plausible as the viewpoint and the body move. The task is difficult because it couples fine-grained appearance transfer (textures, logos, prints) with strict geometric and identity constraints (body shape, pose, self-occlusion, and preservation of the face, hair, and background).

Most research has focused on 2D try-on, which has evolved from warping-and-compositing pipelines built on human parsing and dense correspondence~\cite{han2018viton,choi2021vitonhd,issenhuth2020parserfree} to diffusion models~\cite{ho2020ddpm,song2021ddim,rombach2022ldm} that cast try-on as conditional generation or inpainting~\cite{zhu2023tryondiffusion,kim2024stableviton,choi2024improving,xu2024ootdiffusion,zeng2024catdm}. These models produce highly realistic single images, but they have no mechanism for keeping a garment coherent across viewpoints, which is exactly what a user expects when turning around to see the side or back. Moving to 3D adds two requirements. \emph{Multi-view consistency}: garment details such as patterns, seams, and prints must persist across views without flicker or drift. \emph{Geometric plausibility}: the garment must drape and occlude correctly around the body, including under self-occlusions such as crossed arms or strong torso rotation. The two are tightly coupled: small inconsistencies among edited training images become persistent artifacts in the reconstructed 3D representation, and weak geometric reasoning produces distortions that are most visible in novel views.

In parallel, 3D scene editing has advanced on top of differentiable representations such as Neural Radiance Fields (NeRF)~\cite{mildenhall2020nerf} and 3D Gaussian Splatting (3DGS)~\cite{kerbl2023gaussiansplatting}. The dominant paradigm renders the current 3D scene, edits the renderings with a 2D diffusion model, and distills the edits back into the scene parameters~\cite{haque2023instructnerf2nerf,zhhuang2023dreameditor,chen2023gaussianeditor,wu2024gaussctrl}. Try-on is a demanding instance of this problem: the edit must be \emph{local} (the garment region only), \emph{identity-preserving} (face, hair, body, and background stay fixed), and \emph{faithful to a garment image} rather than to a coarse text prompt, which is why image-prompted 3D editing has become an important direction~\cite{zhhuang2024tipeditor}.

Recent 3DGS-based try-on methods~\cite{chen2024gaussianvton,gsvton2024} adapt this paradigm to garments, but several of their key components are hand-designed rather than learned. (i) Cross-view consistency of the edited training set is encouraged by concatenating attention features from manually selected reference views, rather than by learning correspondences between views. (ii) During 3DGS editing, the diffusion model is conditioned on garment appearance alone and must infer surface orientation, depth ordering, and occlusion boundaries implicitly. (iii) Current-view and reference-view attention features are blended with a fixed weight that ignores camera pose, so every reference view is trusted equally regardless of how close it is to the current viewpoint~\cite{gsvton2024}.

We address these limitations with \textbf{GS-VTON-MVT}, a three-stage framework that combines \emph{implicit} 3D reasoning learned through multi-view attention with \emph{explicit} geometric guidance through multi-modal conditioning. Our guiding principle is that consistency should be \emph{learned} where supervision is strongest (the observed training views) and \emph{enforced} where generalization is hardest (novel views).

\paragraph{(1) Multi-view inpainting transformer.}
Rather than editing each training view separately and reconciling the results post hoc, we edit several views \emph{jointly}. All views are tokenized into a single sequence and processed with cross-view self-attention, so that any patch can attend to every patch in every other view. Multi-view alignment thereby becomes a learnable correspondence problem: attention can discover which regions depict the same garment surface under different viewpoints. The resulting edited set is more coherent and provides a cleaner supervision signal for LoRA personalization~\cite{hu2021lora} and 3D optimization.

\paragraph{(2) Geometry-guided multi-modal ControlNet.}
Cross-view attention is effective for observed views, but the iterative 3DGS editing stage must also synthesize the garment from viewpoints without direct visual evidence, where the diffusion model has to decide how textures warp and where folds form. We therefore condition the editor with a multi-modal ControlNet~\cite{zhang2023controlnet} that receives, in addition to the garment image, depth and surface normals rendered from the current 3DGS state. The garment branch supplies appearance; the normal branch supplies local surface orientation, which governs shading and fold direction; and the depth branch supplies spatial ordering and occlusion boundaries.

\paragraph{(3) Pose-adaptive persona-aware attention.}
Finally, we revisit how reference views are aggregated during 3DGS editing. A fixed blending weight between current-view and reference-view attention implicitly assumes that every reference view is equally informative from every camera pose, which is rarely true. We instead weight reference views by camera proximity (a softmin over pose distances) and predict the overall blending ratio with a lightweight MLP conditioned on the current pose. The editor thus relies on reference evidence when the camera is close to a training viewpoint and on its generative prior when the viewpoint is genuinely novel.

Together, these components form a \emph{hybrid} 3D reasoning scheme: cross-view attention provides implicit 3D awareness when editing training views, and geometry-conditioned diffusion provides explicit structure during novel-view editing. On 3D-VTONBench~\cite{gsvton2024}, GS-VTON-MVT improves multi-view consistency and geometric fidelity over general-purpose 3D editors and prior 3D try-on methods, with the largest gains under self-occlusion and challenging poses.

\paragraph{Contributions.}
In summary:
\begin{itemize}
    \item We propose a \textbf{multi-view inpainting transformer} for constructing the edited training set in 3D try-on, replacing heuristic reference-feature concatenation with cross-view correspondence learned through joint attention.
    \item We introduce a \textbf{geometry-guided multi-modal ControlNet} for 3DGS try-on that extends garment-image conditioning with depth and normal maps rendered from the evolving 3D representation, rather than from a fitted body model.
    \item We develop a \textbf{pose-adaptive persona-aware attention} that weights reference views by camera proximity and predicts the current/reference blending ratio from the camera pose.
    \item We show empirically that combining implicit (attention-based) and explicit (geometry-conditioned) reasoning yields consistent gains in perceptual quality and cross-view consistency for diffusion-based 3D try-on.
\end{itemize}

\section{Related Work}
\label{sec:related_work}

\subsection{Diffusion Models, Personalization, and Controllable Generation}
Denoising diffusion models~\cite{ho2020ddpm} have become the dominant paradigm for high-fidelity image synthesis and editing; faster samplers~\cite{song2021ddim} made them practical for interactive editing, and latent diffusion~\cite{rombach2022ldm} made high-resolution synthesis affordable by denoising in a compact VAE latent space. Vision-language representations such as CLIP~\cite{radford2021clip} ground generation in text or image conditions.
\emph{Personalization} adapts a pretrained model to a specific subject: DreamBooth~\cite{ruiz2023dreambooth} fine-tunes the full model, while LoRA~\cite{hu2021lora} learns low-rank weight updates at a fraction of the cost. \emph{Controllable generation} adds structural constraints; ControlNet~\cite{zhang2023controlnet} attaches a trainable copy of the UNet encoder that injects spatial conditions such as edges, depth, or pose through zero-initialized convolutions. Multiple geometric conditions have been fused for human generation: Champ~\cite{zhu2024champ} conditions human animation on depth, normal, and semantic maps rendered from a parametric body model, and multi-view generators such as MV-Adapter~\cite{huang2025mvadapter} condition on rendered position and normal maps. Our multi-modal ControlNet follows this line but derives depth and normals from the evolving 3DGS scene rather than from a template body or a fixed mesh.

\subsection{2D Image-Based Virtual Try-On}
Classical 2D try-on decomposes the task into human parsing, garment warping, and compositing~\cite{han2018viton}; high-resolution variants reduce misalignment artifacts~\cite{choi2021vitonhd}, and parser-free formulations remove the dependence on brittle segmentation~\cite{issenhuth2020parserfree}. Diffusion-based methods replace explicit warping with conditional generation, producing realistic wrinkles, shading, and fine patterns~\cite{zhu2023tryondiffusion}. Subsequent work improves garment correspondence and identity preservation~\cite{kim2024stableviton,choi2024improving}, controllability and efficiency~\cite{xu2024ootdiffusion,zeng2024catdm}, and architectural simplicity: CatVTON~\cite{chong2025catvton} conditions by simply concatenating person and garment latents, and Leffa~\cite{zhou2025leffa} regularizes attention to behave like a flow field, which sharpens fine garment details. Multi-view try-on has also emerged in 2D: MV-VTON~\cite{wang2025mvvton} uses front and back garment images to dress a person seen from different viewpoints. These models, however, still generate each image independently and provide no guarantee of cross-view coherence, which becomes the main failure mode when their outputs supervise a 3D representation.

\subsection{Multi-View Consistent 3D Editing}
NeRF~\cite{mildenhall2020nerf} and 3DGS~\cite{kerbl2023gaussiansplatting} have enabled diffusion-driven 3D editing, in which renderings are edited in 2D and the edits are distilled back into the scene, either through iterative dataset updates~\cite{haque2023instructnerf2nerf,instructgs2gs2024,dong2024vicanerf} or score distillation~\cite{zhhuang2023dreameditor,chen2023gaussianeditor}; TIP-Editor~\cite{zhhuang2024tipeditor} additionally accepts an image prompt. Because independently edited views disagree, much recent work enforces consistency during 2D editing itself. GaussCtrl~\cite{wu2024gaussctrl} edits all views jointly with a depth-conditioned ControlNet, using depth rendered from the 3DGS, and aligns attention to a few reference views with a fixed blending weight. DGE~\cite{chen2024dge} jointly edits key views with extended self-attention and propagates features to the remaining views along epipolar lines; VcEdit~\cite{wang2024vcedit} alternates 2D editing with 3DGS-based consistency modules; EditSplat~\cite{lee2025editsplat} fuses information from the nearest source views by depth reprojection; and InterGSEdit~\cite{wen2025intergsedit} blends 2D and 3D-consistent attention with a learned gate. A related line performs \emph{multi-view inpainting} directly: NeRFiller~\cite{weber2024nerfiller} inpaints views tiled into grids, MVInpainter~\cite{cao2024mvinpainter} shares reference key/value features across views, and Instant3dit~\cite{barda2025instant3dit} fine-tunes an inpainting model on multi-view grids. More recently, large multi-image editors such as Tinker~\cite{zhao2026tinker} edit several views in a single pass of a diffusion transformer.
Our design borrows from these ideas but targets the specific constraints of try-on: edits are restricted to the garment region, must reproduce a given garment image, and must keep the person's identity intact.

\subsection{3D Virtual Try-On}
Early 3D try-on dressed parametric body models using captured, simulated, or learned garments~\cite{ponsmoll2017clothcap}. Recent methods instead edit real multi-view captures by lifting 2D try-on models into 3D. GaussianVTON~\cite{chen2024gaussianvton} edits a 3DGS scene in multiple stages with a 2D try-on model as an image prompt. GS-VTON~\cite{gsvton2024}, on which we build, personalizes an inpainting diffusion model with LoRA on a reference-driven, jointly edited set of training views, and then edits the 3DGS scene with a persona-aware attention that blends current-view and reference-view features using a fixed weight; it also introduces the 3D-VTONBench benchmark. VTON~360~\cite{he2025vton360} treats 3D try-on as an extension of 2D try-on: it jointly inpaints multi-view renderings of a clothed human using multi-view attention modulated by relative camera pose, conditions on SMPL-X normal maps, and reconstructs the result in 3D.
GS-VTON-MVT differs from these methods in three respects: (i) correspondence between the edited training views is learned with full token-level cross-view attention rather than through single-reference feature sharing; (ii) geometric guidance comes from depth \emph{and} normals rendered from the evolving 3DGS scene, which track the subject's actual silhouette, hair, and loose clothing, rather than from a fitted SMPL-X body; and (iii) the current/reference blending used during 3DGS editing adapts to the camera pose instead of being fixed.

\section{Methodology}\label{sec:method}
\begin{figure}[t]
\centering
\includegraphics[width=\linewidth]{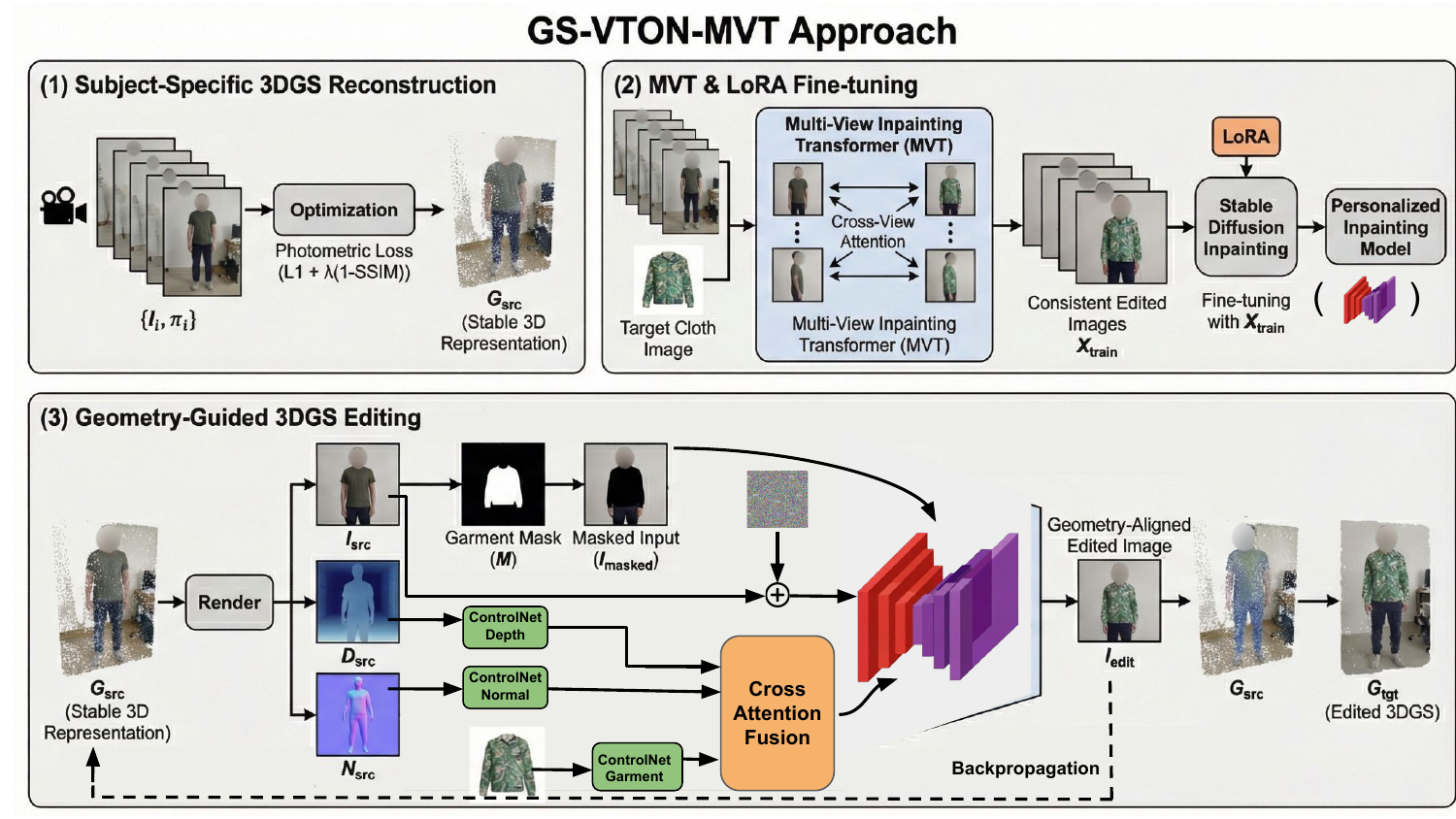}
\caption{\textbf{Overview of GS-VTON-MVT.} \textbf{(1)}~A subject-specific 3DGS scene $\mathcal{G}_{\text{src}}$ is reconstructed from posed multi-view images. \textbf{(2)}~The Multi-View Inpainting Transformer (MVT) jointly edits a subset of training views with cross-view attention, producing a consistent edited set $\mathcal{X}_{\text{train}}$ that is used to personalize a Stable Diffusion inpainting model with LoRA. \textbf{(3)}~During 3DGS editing, RGB, depth, and normal maps are rendered from the current scene; the garment region is masked, and the personalized inpainting model, guided by depth, normal, and garment ControlNet branches fused by cross-attention, produces a geometry-aligned target $I_{\text{edit}}$ that supervises the update of the Gaussians, yielding the edited scene $\mathcal{G}_{\text{tgt}}$.}
\label{fig:overview}
\end{figure}

\subsection{Problem Setup and Overview}\label{sec:method_overview}
Given a multi-view capture of a person
$\mathcal{I}_{\text{train}}=\{(I_i,\pi_i)\}_{i=0}^{m-1}$
with known camera parameters $\pi_i$ (intrinsics and extrinsics) and a garment image $I_{\text{cloth}}$,
our goal is a 3D representation of the person wearing the target garment that renders
consistently from arbitrary viewpoints.
We additionally use per-view clothing-region masks
$\mathcal{M}_{\text{train}}=\{M_i\}_{i=0}^{m-1}$ predicted by an off-the-shelf human parsing model. The masks are not needed for Stage~1 reconstruction; they define the inpainting region in Stage~2 and the editable garment region in Stage~3.

We build on 3D Gaussian Splatting (3DGS)~\cite{kerbl2023gaussiansplatting} and latent-diffusion inpainting~\cite{rombach2022ldm}, and follow the overall structure of GS-VTON~\cite{gsvton2024}, replacing its heuristic components with learned or geometry-aware ones. As illustrated in Fig.~\ref{fig:overview}, \textbf{GS-VTON-MVT} has three stages:
(i) \emph{Stage~1} reconstructs a subject-specific 3DGS scene from the multi-view images (Sec.~\ref{sec:stage1});
(ii) \emph{Stage~2} produces a \emph{consistent} edited training set with a \emph{Multi-View Inpainting Transformer} (MVT) based on cross-view attention~\cite{vaswani2017attention,dosovitskiy2021vit}, which is then used to personalize the inpainting model (Secs.~\ref{sec:stage2}--\ref{sec:lora});
(iii) \emph{Stage~3} edits the 3DGS scene with geometry-guided diffusion, using a \emph{multi-modal ControlNet} (garment + normal + depth)~\cite{zhang2023controlnet} and a \emph{pose-adaptive persona-aware attention} that replaces the fixed view blending of~\cite{gsvton2024} (Secs.~\ref{sec:stage3}--\ref{sec:stage3_obj}).

\subsection{Preliminaries}\label{sec:prelim}

\paragraph{3D Gaussian Splatting (3DGS).}
3DGS represents a scene as a set of anisotropic Gaussians
$\mathcal{G}=\{(\boldsymbol{\mu}_k,\mathbf{\Sigma}_k,\alpha_k,\mathbf{sh}_k)\}_{k=1}^{K}$,
where $\boldsymbol{\mu}_k$ is the 3D mean, $\mathbf{\Sigma}_k$ is the covariance (shape/orientation),
$\alpha_k$ is opacity, and $\mathbf{sh}_k$ are spherical-harmonic appearance coefficients.
Given a camera $\pi$, a differentiable renderer $\mathcal{R}$ produces an RGB image:
\begin{equation}
\hat{I} = \mathcal{R}_{\text{rgb}}(\mathcal{G},\pi),
\end{equation}
using rasterization and alpha compositing as in~\cite{kerbl2023gaussiansplatting}.

\paragraph{Latent diffusion for inpainting.}
We adopt a latent diffusion model (LDM)~\cite{rombach2022ldm} with a VAE encoder/decoder $(E,D)$~\cite{kingma2013vae} and a UNet denoiser~\cite{ronneberger2015unet}.
Given an input image $I$ and mask $M$, we encode $z=E(I)$ and perform masked denoising in latent space:
\begin{equation}
\epsilon_{\theta}(z_t,t,\mathbf{c}) \approx \epsilon,
\end{equation}
where $\mathbf{c}$ denotes conditioning (text/image features).
We use CLIP-based image embeddings for garment conditioning~\cite{radford2021clip} and optionally a caption embedding when text prompting is used (e.g., generated by BLIP-2~\cite{li2023blip2}).

\subsection{Stage 1: Subject-Specific 3DGS Reconstruction}\label{sec:stage1}
We initialize a 3DGS scene $\mathcal{G}_{\text{src}}$ from $\mathcal{I}_{\text{train}}$ using the standard 3DGS optimization~\cite{kerbl2023gaussiansplatting}. The objective minimizes photometric error between rendered views and the observed images:
\begin{equation}
\mathcal{L}_{\text{recon}}(\mathcal{G}) = \sum_{i=0}^{m-1}\left\|\mathcal{R}_{\text{rgb}}(\mathcal{G},\pi_i)-I_i\right\|_1 + \lambda_{\text{ssim}}\left(1-\text{SSIM}(\cdot)\right),
\end{equation}
where $\text{SSIM}$ is computed per view and $\lambda_{\text{ssim}}$ balances terms (following common 3DGS practice~\cite{kerbl2023gaussiansplatting}).
This stage produces geometry/appearance sufficient to render the subject consistently, but \emph{without} the target garment.

\subsection{Stage 2: Multi-View Inpainting Transformer (MVT)}\label{sec:stage2}

\subsubsection{Motivation}
Prior work constructs an edited training set by applying a 2D VTON diffusion model to each view while injecting reference attention features via concatenation in UNet attention blocks~\cite{gsvton2024}.
While helpful, this is still heuristic (manual reference selection, fixed injection structure) and does not explicitly \emph{learn} cross-view correspondences.
We replace this with a learned multi-view transformer that performs \emph{joint} multi-view inpainting via cross-view attention~\cite{vaswani2017attention}.

\subsubsection{Inputs and Tokenization}
We sample $n$ training views (typically $n\in[4,6]$) from $\mathcal{I}_{\text{train}}$:
\begin{equation}
\mathcal{X}=\{(I_i,M_i,\pi_i)\}_{i=0}^{n-1}, \qquad I_{\text{cloth}}.
\end{equation}
Each view image and its mask are concatenated channel-wise and patch-embedded as in vision transformers~\cite{dosovitskiy2021vit}:
\begin{equation}
T_i = \text{PatchEmbed}(\text{Concat}(I_i, M_i)) \in \mathbb{R}^{P\times D},
\end{equation}
where $P$ is the number of patches and $D$ is the embedding dimension.

\subsubsection{Cross-View Self-Attention}
We form a single token sequence by concatenation:
\begin{equation}
T_{\text{all}} = \text{Concat}(T_0,T_1,\dots,T_{n-1}) \in \mathbb{R}^{(nP)\times D}.
\end{equation}
A transformer block applies multi-head self-attention over $T_{\text{all}}$:
\begin{equation}
\text{Attn}(Q,K,V)=\text{softmax}\!\left(\frac{QK^\top}{\sqrt{d_k}}\right)V,
\end{equation}
with $Q=T_{\text{all}}W_Q$, $K=T_{\text{all}}W_K$, $V=T_{\text{all}}W_V$~\cite{vaswani2017attention}.
Because tokens from all views share the same attention space, each patch can attend to potentially corresponding patches in other views, enabling \emph{implicit 3D correspondence learning} from appearance and context.

\subsubsection{Garment Conditioning}
We extract garment tokens using a frozen image encoder (CLIP image tower)~\cite{radford2021clip}:
\begin{equation}
T_{\text{g}}=\text{CLIP}(I_{\text{cloth}})\in\mathbb{R}^{G\times D}.
\end{equation}
We then fuse garment features into the multi-view tokens via cross-attention:
\begin{equation}
T_{\text{fused}}=\text{CrossAttn}(T_{\text{all}}, T_{\text{g}}, T_{\text{g}}).
\end{equation}
Finally we split tokens back into per-view sequences and decode to images:
\begin{equation}
\{T_{\text{out}}^i\}_{i=0}^{n-1}=\text{Split}(T_{\text{fused}}),\qquad
I'_i=\text{Dec}(T_{\text{out}}^i),
\end{equation}
where $\text{Dec}(\cdot)$ is a lightweight CNN decoder (e.g., ResNet-style blocks~\cite{he2016resnet}).

\subsubsection{Stage 2 Objectives}
During pre-training (or teacher-distillation), we supervise MVT outputs against target try-on images $I^{\text{tgt}}_i$ (paired ground truth if available, or pseudo-targets from a strong 2D diffusion VTON model~\cite{choi2024improving}).
We use a combination of reconstruction and perceptual terms:
\begin{equation}
\mathcal{L}_{\text{rec}}=\sum_{i=0}^{n-1}\|I'_i-I^{\text{tgt}}_i\|_1,\qquad
\mathcal{L}_{\text{perc}}=\sum_{i=0}^{n-1}\text{LPIPS}(I'_i,I^{\text{tgt}}_i),
\end{equation}
where LPIPS is a learned perceptual similarity metric~\cite{zhang2018lpips}.
To explicitly regularize cross-view agreement, we add an attention-induced cycle/consistency term:
\begin{equation}
\mathcal{L}_{\text{cyc}}=\sum_{i\neq j}\left\|A_{i\rightarrow j}V_i - V_j\right\|_2,
\end{equation}
where $A_{i\rightarrow j}$ denotes the sub-block of the attention matrix mapping tokens from view $i$ to view $j$.
The full Stage~2 objective is:
\begin{equation}
\mathcal{L}_{\text{stage2}}=\mathcal{L}_{\text{rec}}+\lambda_{\text{perc}}\mathcal{L}_{\text{perc}}+\lambda_{\text{cyc}}\mathcal{L}_{\text{cyc}}.
\end{equation}

\subsection{Personalizing the Inpainting Model via LoRA}\label{sec:lora}
After Stage~2 produces a consistent edited set
$\mathcal{X}_{\text{train}}=\{I'_i\}_{i=0}^{n-1}$,
we personalize a Stable Diffusion inpainting model~\cite{rombach2022ldm,stabilityai2023sd2inpaint} using Low-Rank Adaptation (LoRA)~\cite{hu2021lora}.
Let $\epsilon_{\theta}$ denote the base UNet and $\Delta\theta$ the LoRA parameters; we minimize the standard diffusion noise-prediction objective:
\begin{equation}
\mathcal{L}_{\text{LoRA}}(\Delta\theta)=\mathbb{E}_{\epsilon,t}\Big[\big\|\epsilon-\epsilon_{\theta+\Delta\theta}(z_t,t,\mathbf{c})\big\|_2^2\Big],
\end{equation}
with $z_0=E(I'_i)$ and $z_t=\sqrt{\alpha_t}z_0+\sqrt{1-\alpha_t}\epsilon$.
Following reference-driven completion training strategies, we optionally sample random binary masks during fine-tuning~\cite{tang2024realfill} to improve robustness to mask variation.

\subsection{Stage 3: Geometry-Guided 3DGS Editing with Multi-Modal ControlNet}\label{sec:stage3}

\subsubsection{Multi-Modal Rendering from 3DGS}
At iteration $s$ of 3DGS editing, we render from a camera $\pi$:
\begin{equation}
I_{\text{src}}=\mathcal{R}_{\text{rgb}}(\mathcal{G},\pi),\quad
D_{\text{src}}=\mathcal{R}_{\text{depth}}(\mathcal{G},\pi),\quad
N_{\text{src}}=\mathcal{R}_{\text{normal}}(\mathcal{G},\pi).
\end{equation}
Depth is produced by accumulated expected depth along rays; normals are computed in camera space from depth gradients (finite differences) or attribute rasterization, yielding a dense geometric signal aligned with the rendered viewpoint.

\subsubsection{Multi-Modal ControlNet Conditioning}
ControlNet~\cite{zhang2023controlnet} injects external controls into a diffusion model through a trainable copy of the UNet encoder connected by zero-initialized convolutions. GS-VTON~\cite{gsvton2024} already conditions its personalized inpainting model on the garment image through a ControlNet; we extend this to \emph{three} branches by adding depth and normal conditions:
\begin{equation}
F_{\text{g}}=\text{CNet}_{\text{gar}}(I_{\text{cloth}}),\qquad
F_{\text{n}}=\text{CNet}_{\text{nor}}(N_{\text{src}}),\qquad
F_{\text{d}}=\text{CNet}_{\text{dep}}(D_{\text{src}}).
\end{equation}
We fuse them per-resolution level $\ell$ using either:
\begin{align}
&\textbf{(Concat fusion)} \quad
F_{\text{fused}}^\ell = \text{Conv}_{1\times1}\!\left(\text{Concat}(F_{\text{g}}^\ell,F_{\text{n}}^\ell,F_{\text{d}}^\ell)\right),
\\
&\textbf{(Attn fusion)} \quad
F_{\text{fused}}^\ell=\text{CrossAttn}(F_{\text{g}}^\ell,\text{Concat}(F_{\text{n}}^\ell,F_{\text{d}}^\ell),\text{Concat}(F_{\text{n}}^\ell,F_{\text{d}}^\ell)).
\end{align}
The fused features are injected into the UNet decoder blocks:
\begin{equation}
x_{\text{dec}}^\ell \leftarrow x_{\text{dec}}^\ell + F_{\text{fused}}^\ell.
\end{equation}
Intuitively, garment features specify \emph{appearance}, while normal/depth provide \emph{explicit geometry} (surface orientation, occlusions), which is crucial for unseen viewpoints where implicit multi-view cues are weak.
The depth and normal branches are pretrained ControlNets that we keep frozen, so the geometric guidance requires no additional per-scene training; the garment branch follows GS-VTON~\cite{gsvton2024}. We use the attention fusion in all experiments, as depicted in Fig.~\ref{fig:overview}; the concatenation variant is a lighter alternative.

\subsubsection{Garment-Region Mask and Masked Render}
To perform \emph{inpainting} (rather than global image-to-image translation), we construct a binary mask $M$
for the current rendered view $I_{\text{src}}$ indicating the clothing region to be replaced.
In practice, we apply the same human parsing model used for $\mathcal{M}_{\text{train}}$ to $I_{\text{src}}$
(or equivalently rasterize a projected garment-region label), yielding $M\in\{0,1\}^{H\times W}$ where
$M=1$ denotes editable pixels.

We then form the masked render:
\begin{equation}
I_{\text{masked}} = I_{\text{src}}\odot(1-M),
\end{equation}
which preserves identity/background while exposing only the garment region for synthesis.

\subsubsection{Diffusion-Guided Editing Target}
Let $\epsilon_{\theta+\Delta\theta}$ be the personalized inpainting model.
Conditioned on the masked render, the mask, and multi-modal controls,
the model produces a geometry-aligned edited image:
\begin{equation}
I_{\text{edit}} =
D\!\Big(
\text{Denoise}\big(
E(I_{\text{masked}}),\, M;\, \theta+\Delta\theta,\, F_{\text{fused}}
\big)
\Big),
\end{equation}
where $F_{\text{fused}}=\{F_{\text{fused}}^\ell\}_\ell$ are the fused multi-scale ControlNet features (normal, depth, garment).
The edited image $I_{\text{edit}}$ then serves as the supervision target for updating the 3DGS parameters (Sec.~\ref{sec:stage3_obj}).

\subsection{Pose-Adaptive Persona-Aware Attention}\label{sec:dynamic_attn}
Inside the self-attention layers of the inpainting UNet, GS-VTON~\cite{gsvton2024} blends features computed from the current view with features from reference views using a fixed scalar $\lambda$. A single constant cannot suit every camera: a reference view captured from almost the same pose is highly informative, whereas one seen from the opposite side may be misleading.
We therefore make the blending \emph{pose-adaptive}.

Let $P_{\text{cur}}=(R_{\text{cur}},\mathbf{t}_{\text{cur}})$ denote the current camera extrinsics and $P_i=(R_i,\mathbf{t}_i)$ those of reference training views. We define pose distance:
\begin{equation}
\Delta P_i=\|R_{\text{cur}}-R_i\|_F + \beta\|\mathbf{t}_{\text{cur}}-\mathbf{t}_i\|_2.
\end{equation}
Reference weights are computed by a softmin:
\begin{equation}
w_i=\frac{\exp(-\Delta P_i/\tau)}{\sum_j \exp(-\Delta P_j/\tau)}.
\end{equation}
We then predict a \emph{dynamic} blending factor via a small MLP:
\begin{equation}
\lambda_{\text{dyn}}=\sigma(\text{MLP}(\phi(P_{\text{cur}}))),
\end{equation}
where $\phi(\cdot)$ is a compact pose encoding and $\sigma$ is sigmoid.
Let $F_{\text{cur}}$ denote the attention features computed from the current latent and $F_i$ the corresponding features of reference view $i$ at the same layer and timestep. We aggregate them as
\begin{equation}
F_{\text{ref}}=\sum_i w_i\,F_i,\qquad
F_{\text{final}}=\lambda_{\text{dyn}}F_{\text{cur}}+(1-\lambda_{\text{dyn}})F_{\text{ref}}.
\end{equation}
Near a training viewpoint the softmin concentrates on the closest reference view, favoring consistency with observed evidence; far from all training views the weights flatten, and the MLP can shift weight toward the current-view features, leaving more freedom to the generative prior.
The MLP is lightweight and is optimized separately for each scene.

\subsection{Stage 3 Optimization Objective}\label{sec:stage3_obj}
We update 3DGS parameters by matching the rendered image to the diffusion-edited target while preserving perceptual quality:
\begin{equation}
\mathcal{L}_{\text{img}} =
\lambda_1\big\|I_{\text{src}}-\text{sg}(I_{\text{edit}})\big\|_1
+ \lambda_2\,\text{LPIPS}\!\big(I_{\text{src}},\text{sg}(I_{\text{edit}})\big),
\end{equation}
where $\text{sg}(\cdot)$ denotes stop-gradient on the diffusion output.
Optionally, we also discourage geometric drift with a normal-preservation term:
\begin{equation}
\mathcal{L}_{\text{geo}} = \lambda_3\|N_{\text{src}}-N_{\text{ref}}\|_1,
\end{equation}
where $N_{\text{ref}}$ can be rendered once from the source 3DGS (before editing) to discourage implausible geometry drift.
The full Stage~3 objective is:
\begin{equation}
\mathcal{L}_{\text{stage3}}=\mathcal{L}_{\text{img}}+\mathcal{L}_{\text{geo}}.
\end{equation}

\begin{algorithm}[t]
\caption{GS-VTON-MVT Training/Editing Pipeline}
\label{alg:gsvtonmvt}
\begin{algorithmic}[1]
\State \textbf{Input:} Multi-view images $\mathcal{I}_{\text{train}}$, masks $\mathcal{M}_{\text{train}}$, garment image $I_{\text{cloth}}$
\State \textbf{Stage 1:} Optimize $\mathcal{G}_{\text{src}} \leftarrow \arg\min_{\mathcal{G}} \mathcal{L}_{\text{recon}}(\mathcal{G})$~\cite{kerbl2023gaussiansplatting}
\State \textbf{Stage 2:} Use MVT to jointly edit $n$ views $\Rightarrow \mathcal{X}_{\text{train}}=\{I'_i\}_{i=0}^{n-1}$~\cite{vaswani2017attention,dosovitskiy2021vit}
\State \textbf{Personalize:} Fine-tune SD-inpainting with LoRA on $\mathcal{X}_{\text{train}}$~\cite{rombach2022ldm,hu2021lora}

\For{$s=1$ to $S$}
    \State Sample camera $\pi$; render $(I_{\text{src}},N_{\text{src}},D_{\text{src}})$ from $\mathcal{G}$
    \State Obtain garment mask $M$ on $I_{\text{src}}$; form $I_{\text{masked}}=I_{\text{src}}\odot(1-M)$
    \State Compute pose-adaptive weights $\{w_i\}$ and blend factor $\lambda_{\text{dyn}}$ for $\pi$ (Sec.~\ref{sec:dynamic_attn})
    \State Generate $I_{\text{edit}}$ by inpainting with fused multi-modal ControlNet conditioning~\cite{zhang2023controlnet}
    \State Compute $\mathcal{L}_{\text{stage3}}$ and update $\mathcal{G}$ via backprop through 3DGS~\cite{kerbl2023gaussiansplatting}
\EndFor

\State \textbf{Output:} Edited 3DGS model $\mathcal{G}_{\text{tgt}}$ for novel-view virtual try-on
\end{algorithmic}
\end{algorithm}

\section{Experiments}\label{sec:experiments}

\subsection{Dataset and Evaluation Protocol}\label{sec:protocol}
We evaluate on \textbf{3D-VTONBench}~\cite{gsvton2024}, a multi-view benchmark for 3D try-on containing $60$ subjects captured in a variety of poses and garments. In our setup each subject has $12$--$16$ views, which we process at $1024\times 1024$ resolution.
For each subject, we select $n{=}4$ training views for Stage~2 multi-view editing and hold out the remaining views for novel-view evaluation. During Stage~3, we optimize the 3DGS representation~\cite{kerbl2023gaussiansplatting} with cameras sampled along the capture trajectory.

\subsection{Metrics}\label{sec:metrics}
We report perceptual, consistency, and geometric metrics.
(i)~\textbf{FID}~\cite{heusel2017fid} and \textbf{KID}~\cite{binkowski2018kid} measure distributional realism of the rendered views.
(ii)~\textbf{LPIPS}~\cite{zhang2018lpips} between renderings of random view pairs of the same subject and garment serves as a proxy for multi-view consistency; lower values indicate fewer view-dependent changes.
(iii)~\textbf{PSNR} is computed on held-out camera views against the original capture and is restricted to pixels outside the garment mask, because no ground-truth image of the subject wearing the target garment exists; it therefore measures how well identity, background, and unedited clothing are preserved in novel views.
(iv)~\textbf{Normal MSE} compares normal maps rendered from the edited scene with normals rendered from the pre-edit 3DGS scene, again outside the garment mask, and measures geometric drift in regions that should not change.

\subsection{Baselines}\label{sec:baselines}
We compare against the 3D try-on method GS-VTON~\cite{gsvton2024} and the general-purpose 3D editors used in its evaluation: Instruct-NeRF2NeRF~\cite{haque2023instructnerf2nerf}, Instruct-GS2GS~\cite{instructgs2gs2024}, GaussianEditor~\cite{chen2023gaussianeditor}, GaussCtrl~\cite{wu2024gaussctrl}, and ViCA-NeRF~\cite{dong2024vicanerf}.
All methods receive the same garment image and camera splits; methods not designed for try-on are restricted to the clothing-region mask and given consistent prompt and conditioning templates.
More recent multi-view consistent editors (DGE~\cite{chen2024dge}, EditSplat~\cite{lee2025editsplat}) and the multi-view try-on method VTON~360~\cite{he2025vton360} are not yet included in this comparison; we discuss them in Sec.~\ref{sec:conclusion}.
Neither GS-VTON nor the general-purpose editors report FID, KID, or LPIPS on 3D-VTONBench. All baseline numbers in Tables~\ref{tab:sota} and~\ref{tab:ablation} are therefore obtained by running the public implementations ourselves on the same subjects, garments, and camera splits as our method.

\subsection{Implementation Details}\label{sec:impl}
All views are processed at $1024\times1024$ resolution. Stage~2 jointly edits $n{=}4$ training views per subject, with pseudo-targets for training the MVT produced by a 2D diffusion try-on model~\cite{choi2024improving}. The edited views are used to personalize a Stable Diffusion inpainting model~\cite{stabilityai2023sd2inpaint} with LoRA~\cite{hu2021lora}. In Stage~3, the depth and normal ControlNet branches are pretrained and frozen, the garment branch follows GS-VTON~\cite{gsvton2024}, the three branches are combined with the attention fusion, and the blending MLP of Sec.~\ref{sec:dynamic_attn} is optimized per scene. The same off-the-shelf human parsing model provides the clothing masks for the training views in Stage~2 and for the rendered views in Stage~3. Stage~3 uses the same number of optimization iterations as GS-VTON.

\subsection{Quantitative Comparison}\label{sec:quant}
Table~\ref{tab:sota} reports the main results. GS-VTON-MVT obtains the best FID, KID, and LPIPS among all methods. Relative to general-purpose 3D editors, we attribute the gains to (i)~learned cross-view correspondence in Stage~2 and (ii)~explicit depth and normal conditioning in Stage~3.

\begin{table*}[t]
\centering
\caption{Comparison with prior methods on 3D-VTONBench. Lower is better for all metrics.}
\label{tab:sota}
\begin{tabular}{lccc}
\toprule
Method & FID$\downarrow$ & LPIPS$\downarrow$ & KID$\downarrow$ \\
\midrule
Instruct-NeRF2NeRF~\cite{haque2023instructnerf2nerf} & 52.3 & 0.231 & 0.041 \\
Instruct-GS2GS~\cite{instructgs2gs2024}              & 47.8 & 0.198 & 0.035 \\
GaussianEditor~\cite{chen2023gaussianeditor}         & 44.2 & 0.176 & 0.029 \\
GaussCtrl~\cite{wu2024gaussctrl}                     & 41.5 & 0.165 & 0.026 \\
ViCA-NeRF~\cite{dong2024vicanerf}                    & 43.1 & 0.171 & 0.028 \\
GS-VTON~\cite{gsvton2024}                            & 38.2 & 0.142 & 0.023 \\
\midrule
\textbf{GS-VTON-MVT (Ours)}                          & \textbf{24.1} & \textbf{0.076} & \textbf{0.008} \\
\bottomrule
\end{tabular}
\end{table*}

\paragraph{Novel-view fidelity and geometry.}
On held-out views, GS-VTON-MVT raises PSNR outside the garment region from $21.3$\,dB (GS-VTON) to $26.2$\,dB, indicating that identity and background are better preserved in novel views, and reduces the normal-map MSE from $0.087$ to $0.041$, indicating less geometric drift in unedited regions, consistent with the role of explicit depth and normal conditioning during 3DGS optimization.

\subsection{Ablation Studies}\label{sec:ablation}
Table~\ref{tab:ablation} isolates the contribution of each component, starting from GS-VTON (A) and adding the components individually (B--D) and in combination (E--F). The multi-view inpainting transformer yields the largest single-component gain (B), supporting the claim that learned cross-view attention enforces multi-view coherence better than single-reference feature concatenation. The multi-modal ControlNet (C) gives a smaller but complementary improvement, which we observe mainly as better geometric plausibility under self-occlusion and in unseen viewpoints. Pose-adaptive attention alone (D) has the smallest effect, but adding it to the other two components (E$\rightarrow$F) still improves every metric, with the clearest benefit on viewpoints far from the training cameras.

\begin{table}[t]
\centering
\caption{Ablation of GS-VTON-MVT components on 3D-VTONBench.}
\label{tab:ablation}
\begin{tabular}{lccc}
\toprule
Variant & FID$\downarrow$ & LPIPS$\downarrow$ & KID$\downarrow$ \\
\midrule
A: Baseline (GS-VTON)~\cite{gsvton2024} & 38.2 & 0.142 & 0.023 \\
B: A + MVT                           & 32.1 & 0.118 & 0.017 \\
C: A + Multi-modal ControlNet        & 34.5 & 0.125 & 0.019 \\
D: A + Pose-adaptive attention       & 36.8 & 0.135 & 0.021 \\
E: A + MVT + Multi-modal ControlNet  & 27.3 & 0.092 & 0.011 \\
F: \textbf{Full model}               & \textbf{24.1} & \textbf{0.076} & \textbf{0.008} \\
\bottomrule
\end{tabular}
\end{table}

\paragraph{Consistency versus geometry.}
The multi-view inpainting transformer mainly reduces cross-view inconsistency (variant~B in Table~\ref{tab:ablation}), whereas the multi-modal ControlNet mainly reduces the normal-map error (0.087 $\rightarrow$ 0.053). The full model combines both effects (LPIPS 0.076, normal MSE 0.041), supporting the hybrid hypothesis: implicit correspondence learning matters most for the observed views, and explicit geometric guidance matters most for novel views.

\subsection{Qualitative Results}\label{sec:qual}
\begin{figure}[t]
\centering
\footnotesize
\makebox[\linewidth][l]{%
\makebox[0.2790\linewidth]{Input views}\hspace{0.0083\linewidth}%
\makebox[0.1381\linewidth]{Garment}\hspace{0.0083\linewidth}%
\makebox[0.2790\linewidth]{\textbf{Ours}}\hspace{0.0083\linewidth}%
\makebox[0.2790\linewidth]{\qualbaseline}}\\[2pt]
\includegraphics[width=\linewidth]{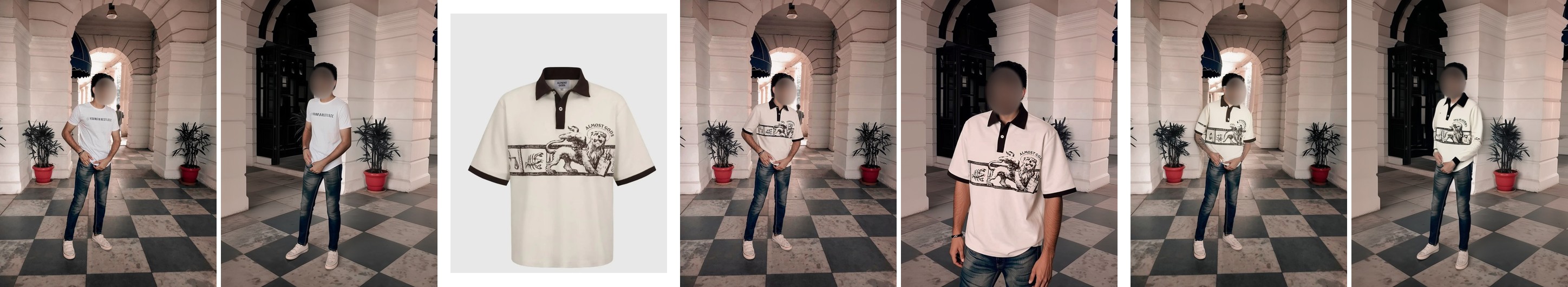}\\[4pt]
\includegraphics[width=\linewidth]{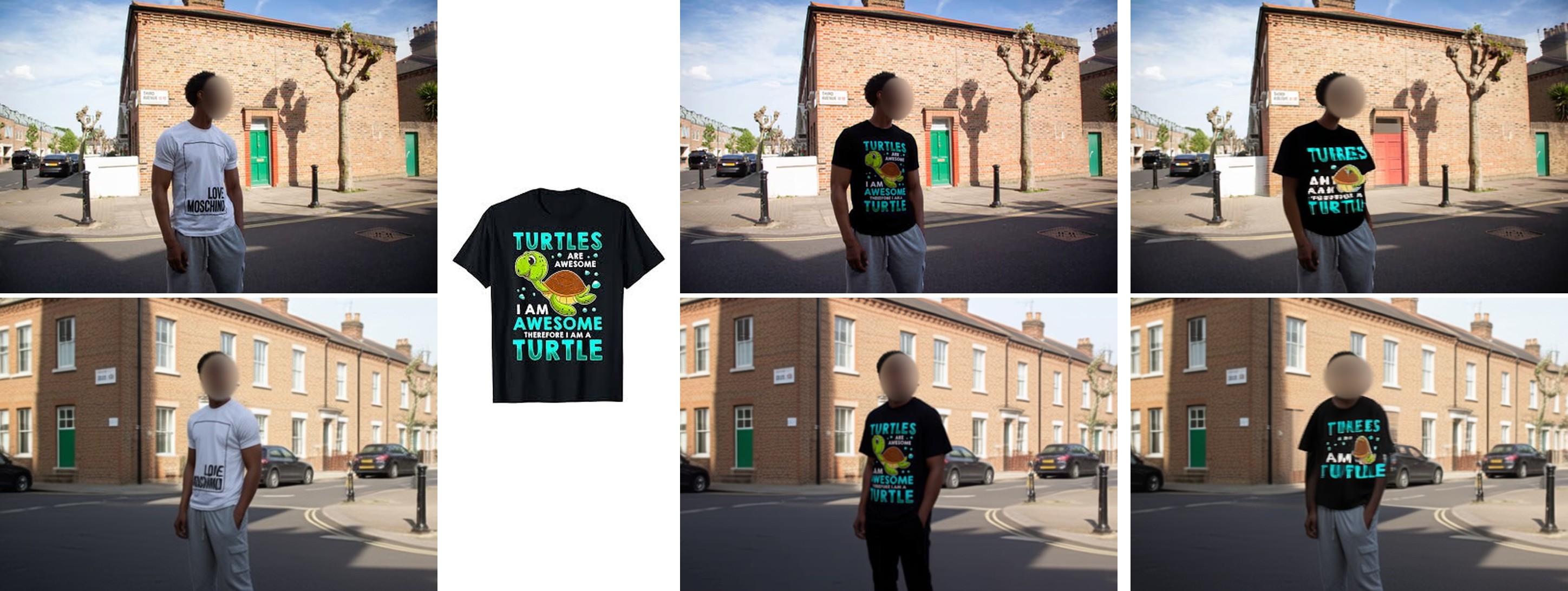}\\[4pt]
\includegraphics[width=\linewidth]{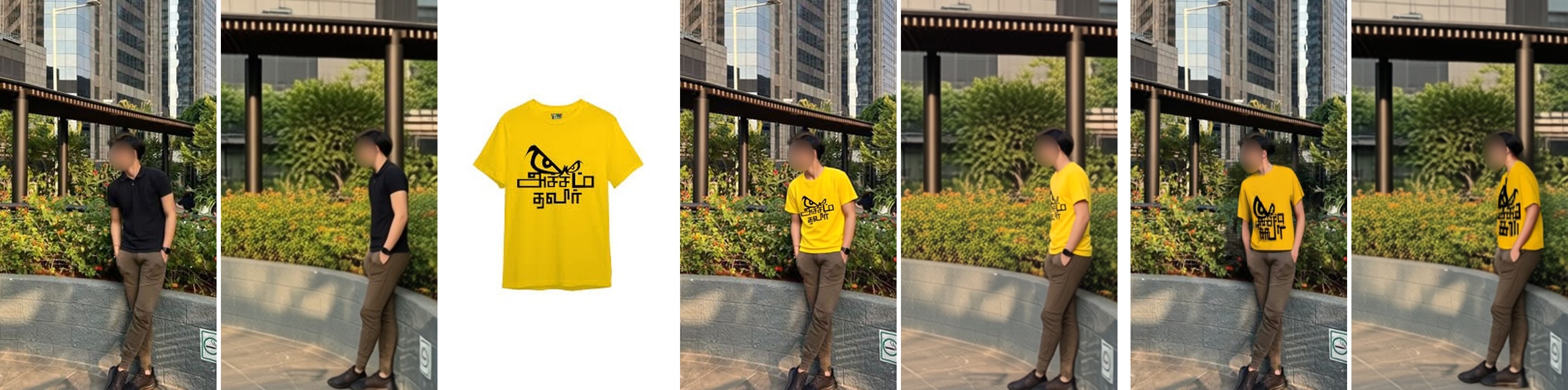}
\caption{\textbf{Qualitative comparison on in-the-wild subjects.} Each subject is captured from multiple views and processed with the full pipeline; we show two of the input views, the target garment, and the same two views produced by GS-VTON-MVT and by the baseline. Baseline outputs are resized to the aspect ratio of the input. Faces are blurred for privacy.}
\label{fig:qual}
\end{figure}

Figure~\ref{fig:qual} shows try-on results for three in-the-wild subjects outside the benchmark. For each subject we run the full three-stage pipeline on a multi-view capture, exactly as for the benchmark subjects, and show two of the views together with the corresponding outputs of the baseline. GS-VTON-MVT reproduces the print of the target garment in both views and leaves the background and the remaining clothing of the subject unchanged. The baseline alters parts of the scene outside the garment (for example, the door behind the second subject changes color) and distorts printed text.
More generally, we observe three recurring failure modes of prior methods that GS-VTON-MVT mitigates.
\textbf{(1) Pattern integrity:} baselines often smear high-frequency textures such as stripes, logos, and printed text across views, whereas cross-view attention keeps patch-level correspondences stable.
\textbf{(2) Self-occlusion:} for crossed arms or hands in pockets, depth conditioning preserves the correct front-to-back ordering and reduces hallucinated garment layers.
\textbf{(3) Drape and shading:} normal conditioning improves fold orientation and avoids a flat, ``stuck-on'' appearance, producing wrinkles that stay coherent across nearby viewpoints.

\subsection{User Study}\label{sec:userstudy}
We conducted a user study with $25$ participants, following the protocol of GS-VTON~\cite{gsvton2024}. Each participant evaluated $25$ randomly sampled subject--garment pairs. For each pair, the results of all seven methods were shown side by side and the participant selected the best one with respect to \textit{realism}, \textit{garment similarity}, and \textit{overall quality}, giving $625$ responses per criterion. Table~\ref{tab:userstudy} reports the share of responses in which each method was selected. GS-VTON-MVT is preferred most often on all three criteria.

\begin{table}[t]
\centering
\caption{User study: preference rate (\%) per method for each criterion. Higher is better.}
\label{tab:userstudy}
\begin{tabular}{lccc}
\toprule
Method & Realism$\uparrow$ & Garment sim.$\uparrow$ & Overall$\uparrow$ \\
\midrule
Instruct-NeRF2NeRF~\cite{haque2023instructnerf2nerf} & 8  & 6  & 7  \\
Instruct-GS2GS~\cite{instructgs2gs2024}              & 12 & 10 & 11 \\
GaussianEditor~\cite{chen2023gaussianeditor}         & 10 & 12 & 11 \\
GaussCtrl~\cite{wu2024gaussctrl}                     & 14 & 16 & 15 \\
ViCA-NeRF~\cite{dong2024vicanerf}                    & 9  & 8  & 8  \\
GS-VTON~\cite{gsvton2024}                            & 18 & 20 & 19 \\
\midrule
\textbf{GS-VTON-MVT (Ours)}                          & \textbf{29} & \textbf{28} & \textbf{29} \\
\bottomrule
\end{tabular}
\end{table}

\subsection{Efficiency}\label{sec:efficiency}
Stage~2 is run once per subject--garment pair to produce the edited training set, and Stage~3 uses the same number of optimization iterations as GS-VTON. The complete pipeline takes about $75$ minutes per subject--garment pair.

\section{Conclusion}\label{sec:conclusion}
We presented \textbf{GS-VTON-MVT}, a diffusion-based 3D virtual try-on framework that improves multi-view consistency and geometric fidelity by combining implicit and explicit 3D reasoning. Building on GS-VTON~\cite{gsvton2024}, we replace its single-reference feature concatenation with a multi-view inpainting transformer that learns cross-view correspondence among the observed views, add depth and normal maps rendered from the evolving 3DGS scene as ControlNet conditions alongside the garment image, and replace the fixed current/reference attention blending with a pose-adaptive scheme. On 3D-VTONBench, these changes improve FID, KID, LPIPS-based consistency, and user preference. The results support our central hypothesis: implicit cross-view reasoning is most effective for keeping observed views coherent, whereas explicit geometric conditioning is most valuable for stable novel-view synthesis.

\paragraph{Limitations and future work.}
Several limitations remain.
\emph{Geometry of the original garment.} The depth and normal conditions are rendered from the current 3DGS scene, which initially encodes the subject's \emph{original} clothing; when the target garment has a different silhouette (\eg, long sleeves replacing short ones, or a dress replacing trousers), this guidance can bias the edit toward the old shape, and the normal-preservation term $\mathcal{L}_{\text{geo}}$ can likewise resist legitimate geometric change. Restricting $\mathcal{L}_{\text{geo}}$ to regions outside the garment mask, dilating the mask, or blending in body-model geometry are natural remedies.
\emph{Noisy rendered geometry.} Depth and normals rendered from 3DGS are noisy, particularly at thin structures and hair; more geometrically accurate Gaussian representations could make the guidance more reliable.
\emph{Scalability.} Full cross-view attention is quadratic in the number of views, which currently limits Stage~2 to a handful of views; sparse, epipolar, or factorized attention could relax this.
\emph{Hard cases.} Long hair crossing the garment and severe self-occlusion still produce occasional artifacts, suggesting the need for better human parsing and stronger occlusion priors.
\emph{Evaluation.} Our comparison does not yet include recent multi-view consistent editors~\cite{chen2024dge,lee2025editsplat} or the multi-view try-on method VTON~360~\cite{he2025vton360}, and our consistency metric is LPIPS-based; evaluating with dedicated multi-view consistency metrics such as MEt3R~\cite{asim2025met3r} and with garment-fidelity measures computed on the garment region is an important next step.
Finally, extending the framework to layered outfits, long coats, and dynamic (video) try-on is a promising direction.

\bibliographystyle{splncs04}
\bibliography{main}
\end{document}